\documentclass[10pt,twocolumn,letterpaper]{article}

\usepackage[pagenumbers]{cvpr}

\definecolor{cvprblue}{rgb}{0.21,0.49,0.74}
\usepackage[pagebackref,breaklinks,colorlinks,allcolors=cvprblue]{hyperref}

\def\paperID{***}
\def\confName{CVPR}
\def\confYear{2025}
\usepackage[accsupp]{axessibility}
\usepackage[utf8]{inputenc}
\usepackage[T1]{fontenc}
\usepackage{hyperref}
\usepackage{url}
\usepackage{booktabs}
\usepackage{amsfonts}
\usepackage{nicefrac}
\usepackage{microtype}
\usepackage{xcolor}

\usepackage{microtype}
\usepackage{lipsum}
\usepackage{graphicx}
\usepackage{subcaption}
\usepackage{sidecap}
\usepackage{caption}
\usepackage{booktabs} %
\usepackage{amsfonts}       %
\usepackage{nicefrac}       %
\usepackage{microtype}      %
\usepackage{comment}
\usepackage{enumitem}
\usepackage{wrapfig,tikz}
\usepackage{amsmath,longtable,fancyhdr}
\usepackage{bigstrut, tabularx, multirow, makecell, diagbox}
\usepackage[export]{adjustbox}
\usepackage{graphicx,float}
\usepackage{amssymb,amsthm, mathtools}
\usepackage{stackrel}
\usepackage{color}
\usepackage{colortbl}
\usepackage{theoremref}
\usepackage{cases}
\usepackage{stmaryrd}
\usepackage{mathabx}
\usepackage{dsfont}
\usepackage{xcolor}
\usepackage{algorithmic}
\usepackage{algorithm}
\usepackage{bm}

\newcolumntype{C}[1]{>{\centering\arraybackslash}m{#1}}
\newcolumntype{P}[1]{>{\centering\arraybackslash}p{#1}}

\usepackage{mdframed}

\makeatletter
\usepackage{xspace} 
\def\@onedot{\ifx\@let@token.\else.\null\fi\xspace}
\DeclareRobustCommand\onedot{\futurelet\@let@token\@onedot}

\usepackage{adjustbox}
\usepackage{amsmath,amsfonts,bm}

\def\eqref#1{equation~\ref{#1}}

\def\1{\bm{1}}

\def\bfa{{\mathbf{a}}}

\def\bfh{{\mathbf{h}}}

\def\bfp{{\mathbf{p}}}
\def\bfq{{\mathbf{q}}}
\def\bfr{{\mathbf{r}}}

\def\bfv{{\mathbf{v}}}
\def\bfw{{\mathbf{w}}}
\def\bfx{{\mathbf{x}}}
\def\bfy{{\mathbf{y}}}

\DeclareMathAlphabet{\mathsfit}{\encodingdefault}{\sfdefault}{m}{sl}
\SetMathAlphabet{\mathsfit}{bold}{\encodingdefault}{\sfdefault}{bx}{n}

\def\gD{{\mathcal{D}}}

\def\gF{{\mathcal{F}}}

\def\gH{{\mathcal{H}}}

\def\gL{{\mathcal{L}}}

\def\gT{{\mathcal{T}}}

\def\bfA{{\mathbf{A}}}

\def\bfF{{\mathbf{F}}}

\def\bfJ{{\mathbf{J}}}

\def\bfP{{\mathbf{P}}}

\def\bfT{{\mathbf{T}}}

\def\bfV{{\mathbf{V}}}
\def\bfW{{\mathbf{W}}}

\newcommand{\R}{\mathbb{R}}

\DeclareMathOperator*{\argmin}{arg\,min}

\newcommand{\x}{\mathbf{x}}
\newcommand{\xadv}{\mathbf{x}^{adv}}
\newcommand{\xs}{\mathbf{x}^{s}}
\newcommand{\xt}{\mathbf{x}^{t}}

\newcommand{\w}{\mathbf{w}}

\newcommand{\e}{\mathbf{e}}

\title{Improving the Transferability of Adversarial Attacks on Face Recognition with Diverse Parameters Augmentation}
\author{Fengfan Zhou$^{1}$, 
Bangjie Yin$^{2}$,
Hefei Ling$^1$\thanks{\textit{Corresponding author.}} ,
Qianyu Zhou$^{3}$, 
Wenxuan Wang$^{4}$
\\$^1$School of Computer Science and Technology, Huazhong University of Science and Technology; \\$^2$Shanghai Shizhuang Information Technology Co., Ltd; \\$^3$ Department of Computer Science and Engineering, Shanghai Jiao Tong University; 
\\$^4$ School of Computer Science, Northwestern Polytechnical University.\\
$^1${\tt\small \{ffzhou, lhefei\}@hust.edu.cn}, $^2${\tt\small jamesyin10@gmail.com},  \\ $^3${\tt\small
zhouqianyu@sjtu.edu.cn}, $^4${\tt\small wxwang@nwpu.edu.cn }
}

\begin{document}
\maketitle
\begin{abstract}
    Face Recognition (FR) models are vulnerable to adversarial examples that subtly manipulate benign face images, underscoring the urgent need to improve the transferability of adversarial attacks in order to expose the blind spots of these systems. Existing adversarial attack methods often overlook the potential benefits of augmenting the surrogate model with diverse initializations, which limits the transferability of the generated adversarial examples. To address this gap, we propose a novel method called Diverse Parameters Augmentation (DPA) attack method, which enhances surrogate models by incorporating diverse parameter initializations, resulting in a broader and more diverse set of surrogate models. Specifically, DPA consists of two key stages: Diverse Parameters Optimization (DPO) and Hard Model Aggregation (HMA). In the DPO stage, we initialize the parameters of the surrogate model using both pre-trained and random parameters. Subsequently, we save the models in the intermediate training process to obtain a diverse set of surrogate models. During the HMA stage, we enhance the feature maps of the diversified surrogate models by incorporating beneficial perturbations, thereby further improving the transferability. Experimental results demonstrate that our proposed attack method can effectively enhance the transferability of the crafted adversarial face examples.
\end{abstract}
\vspace{-0.5cm}    
\section{Introduction}
\begin{figure}[htbp]
	\begin{center}
		\centerline{\includegraphics[width=80mm]{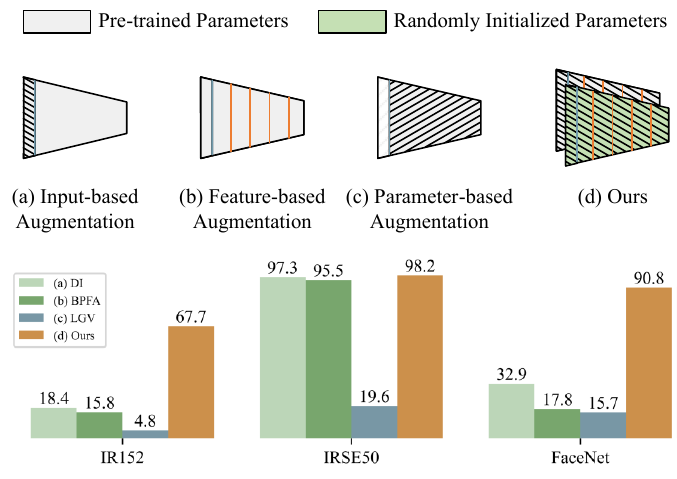}}
		\caption{
        Top: comparison between traditional augmentation-based adversarial attack methods and our proposed method. The black pattern filling on the left and right sides of the blue line represents input-based and parameter-based augmentation, respectively. The orange pattern filling indicates feature-based augmentation. Bottom: comparison of performance among 4 types of augmentations.
		}
		\label{fig:motivation}
	\end{center}
        \vspace{-0.8cm}
\end{figure}
Owing to the relentless progress in deep learning, Face Recognition (FR) has achieved substantial advancements \cite{facenet, cosface, arcface, partial_fc, idiff_face, DBLP:conf/cvpr/LiGLHBYYS23}.
However, the inherent vulnerabilities in existing FR models pose significant security risks \cite{das2021towards, amel, zhou2022generative, iadg, narayan2023df, wang2024disentangle, zhou2024test}, with adversarial attacks emerging as one of the critical concerns.
Therefore, there is an urgent need to bolster the resilience of FR models against adversarial face examples to reveal and address the vulnerabilities.

As a result, numerous research efforts have been focused on this area. Several adversarial attacks have been developed to create adversarial face examples with features like stealthiness \cite{semantic_adv, tip_im, low_key, amt_gan, clip2protect}, transferability \cite{dfanet, sibling_attack, bpfa, adv_restore}, and the capacity for physical attacks \cite{adv_makeup, at3d, cvpr23_opt_attack}. These initiatives are aimed at improving the effectiveness of adversarial attacks on FR. Nevertheless, the transferability of these adversarial attacks remains limited.
To enhance the transferability of adversarial attacks, augmentation emerges as one of the most effective methods. As illustrated in \cref{fig:motivation}, augmentation-based adversarial attack consists of three augmentation types: input-based \cite{dim, ssa6, sia, bsr}, feature-based \cite{dfanet, bpfa}, and parameter-based \cite{lgv, sasd}.

Most augmentation-based adversarial attack methods focus on input-based augmentation to improve transferability. Previous research highlights a symmetry between surrogate models used in adversarial attacks and input data in training tasks \cite{mim, ni_fgsm, bpfa}. In training, data augmentation has proven effective in enhancing model generalization. Drawing on this symmetry, augmenting models for crafting adversarial examples can yield examples with greater transferability. Input-based augmentation can be seen as methods that only augment models at the input layer, demonstrating significant effectiveness in enhancing transferability. However, augmenting surrogate models in deeper layers (\ie, feature-based and parameter-based augmentation) offers a more direct form of augmentation. Nevertheless, few studies explore deep-layer augmentation \cite{bpfa, lgv, sasd}. As a typical parameter-based augmentation method in deep layers, LGV \cite{lgv} uses a pre-trained surrogate model and collects multiple parameter sets through additional training epochs with a high, constant learning rate, thereby enhancing the transferability of adversarial examples. Although these parameter-based augmentation methods \cite{lgv, sasd} show promising effectiveness, they face two problems: (1) \textit{Static surrogate model initializations}: these methods solely augment the surrogate model from pre-trained parameters, limiting the parameter diversity of the surrogate models and thereby hindering the transferability of the crafted adversarial examples. (2) \textit{Unavailability of the FR head}: modern FR training procedures typically involve training both a backbone model and a head model \cite{cosface, arcface}.
Following training, inference is conducted solely with the backbone models, while the head models are often not released as open-source. Consequently, in the majority of instances, we are unable to access the pre-trained parameters for the head models. Traditional parameter-based augmentation adversarial attacks augment models from pre-trained parameters \cite{lgv, sasd}. This limitation makes it challenging to craft adversarial examples on FR models with numerous open-sourced FR models.

To address these problems, we introduce a novel adversarial attack called Diverse Parameters Augmentation (DPA) to enhance the transferability of crafted adversarial face examples. Unlike existing parameter-based augmentation adversarial attack methods that overlook the use of diverse parameter initializations \cite{lgv, sasd}, we diversify parameter initializations with both random and pre-trained values, thereby improving black-box attack capacity.

Technically, our proposed attack method comprises two stages: Diverse Parameters Optimization (DPO) and Hard Model Aggregation (HMA).
During the DPO stage, we initiate a subset of the optimization parameters with random noise, preserving the refined parameters in the intermediate training process to yield a diverse set of surrogate models, which is instrumental in bolstering the transferability of the adversarial examples.
In the HMA stage, we add beneficial perturbations \cite{Beneficial_Perturbation_Network} with optimization directions opposite to those of the adversarial perturbations onto the feature maps of the parameter-augmented surrogate models, achieving the effect of hard model augmentation \cite{bpfa} to further enhances the transferability of the crafted adversarial examples. Our proposed attack method effectively addresses the challenges posed by the lack of diverse parameters and the absence of the FR head. The comparison between traditional adversarial attack methods and our proposed method is illustrated in \cref{fig:motivation}.
By using diverse initializations for surrogate models, we can expand the parameter set of these models, thereby enhancing the transferability of the adversarial examples.

Our main contributions are summarized as follows:
\begin{itemize}
   \item We introduce a novel perspective that parameter-based augmentation adversarial attacks should augment the parameters of surrogate models by incorporating diverse initialized parameters. To the best of our knowledge, this is the first adversarial attack on FR that utilizes parameter augmentation to enhance transferability.
   \item  We introduce a new adversarial attack method on FR, called DPA, which comprises DPO and HMA stages. The DPO stage uses both pre-trained and random initializations to optimize the model and save the models in the intermediate training process to diversify the surrogate model set. The HMA stage adds beneficial perturbations with optimization directions opposite to those of adversarial perturbations onto the feature maps of the diversified surrogate models to further enhance transferability.
   \item Extensive experiments reveal that our proposed method attains superior performance when compared with the state-of-the-art adversarial attack methods.
\end{itemize}
\begin{figure*}[htbp]
	\begin{center}
		\centerline{\includegraphics[width=170mm]{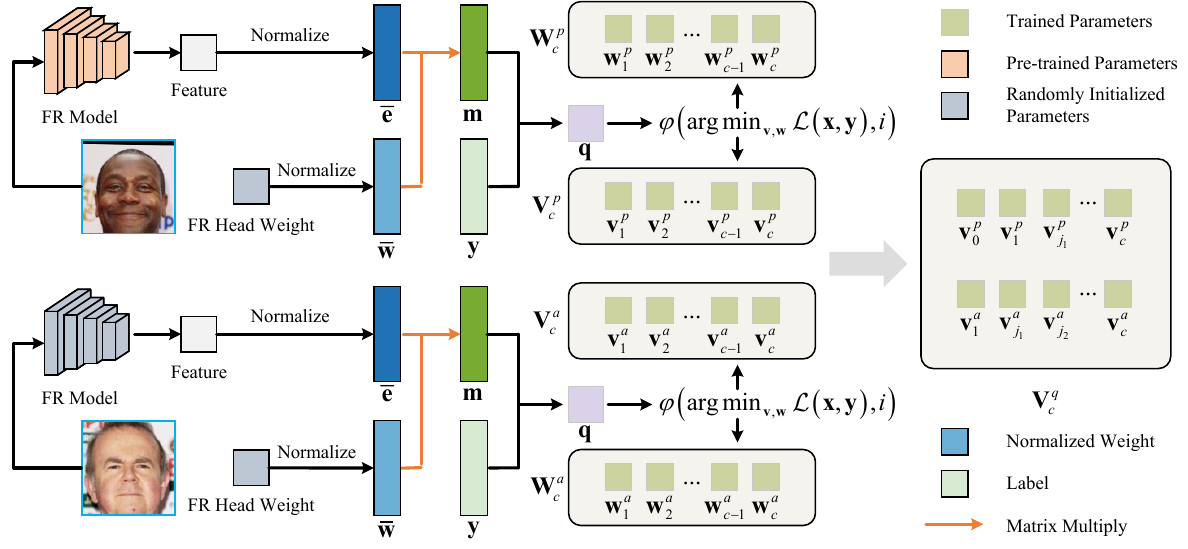}}
		\caption{The framework of the Diverse Parameters Optimization (DPO). We enhance the diversity of the surrogate model parameters by integrating both pre-trained and random initializations. The method yields a diverse set of surrogate model parameters, which enhances the parameter diversity of the surrogate FR models and consequently improves transferability of the crafted adversarial examples. 
		}
		\label{fig:dpo}
	\end{center}
    \vspace{-0.8cm}
\end{figure*}
\section{Related Work}
\noindent \textbf{Adversarial Attacks.}
The primary objective of adversarial attacks is to introduce subtle perturbations into benign images, thereby deceiving machine learning systems and inducing them to produce incorrect predictions \cite{ax_init, fgsm}. The presence of adversarial examples constitutes a substantial security threat to contemporary machine learning systems. Consequently, significant research efforts have been expended to investigate adversarial attacks, with the aim of bolstering system robustness \cite{ssa6, liang2023adversarial, lu2023set, shayegani2024jailbreak, ge2023improving, zhou2023advclip, zhang2022towards, mingxing2021towards}. To enhance the potency of black-box adversarial attacks, DI \cite{dim} incorporates random transformations into adversarial examples at each iteration, effectively achieving data augmentation. VMI-FGSM \cite{vt} harnesses gradient variance to stabilize the update process, thereby enhancing black-box attack performance. SSA \cite{ssa6} translates adversarial examples into the frequency domain and applies spectral manipulation for augmentation. SIA \cite{sia} introduces random image transformations on each image block, producing diverse variations for gradient estimation. BSR \cite{bsr} segments the input image into multiple blocks, randomly shuffling and rotating them to generate a set of new images for gradient computation.
Despite these advancements, these methods overlook the potential benefit of augmenting the surrogate model with diverse initializations, which limits the transferability of the crafted adversarial examples. In contrast, our proposed method augments surrogate models using both pre-trained and randomly initialized parameters, producing a broader and more diverse set of surrogate models.

\noindent \textbf{Adversarial Attacks on Face Recognition.}
Adversarial attacks on FR models can be classified based on the constraints imposed on the adversarial perturbations. These attacks are broadly categorized into two types: restricted attacks \cite{DBLP:conf/iccv/LiuFWYM23, DBLP:conf/iccv/XuGZCMW23, mim, DBLP:conf/iccv/ZhuRSYJ23, lu2023set, DBLP:conf/iccv/LiuZZQD23, DBLP:conf/iccv/ChenYCCL23, rma} and unrestricted attacks \cite{DBLP:conf/nips/ChenLWJDZ23, DBLP:conf/nips/YuanZGCS22, DBLP:journals/corr/abs-2307-12499, DBLP:conf/nips/StimbergCL0SQLK23, DBLP:conf/iccv/SuryantoKLKLHYO23, DBLP:conf/iccv/WangYJLC23, DBLP:conf/iccv/WeiHS023, DBLP:conf/cvpr/WeiYH23}. Restricted attacks create adversarial examples that adhere to a specified perturbation limit, such as an $L_p$ norm bound.
Our proposed attack method is a restricted attack. Consequently, we will delve into the specifics of restricted attacks in detail.
To enhance the transferability of adversarial attacks on FR, Zhong and Deng \cite{dfanet} introduced DFANet, which employs dropout on the feature maps of convolutional layers to achieve an ensemble-like effect. Zhou et al. \cite{bpfa} proposed BPFA, enhancing attack transferability by integrating beneficial perturbations \cite{Beneficial_Perturbation_Network} onto the feature maps of FR models, resulting in the effect of hard model augmentation. Li et al. \cite{sibling_attack} leveraged additional information from FR-related tasks and applied a multi-task optimization framework to further improve the transferability of adversarial examples.
Unrestricted adversarial attacks conversely generate adversarial examples without the limitations of a predefined perturbation bound. These attacks are primarily focused on physical attacks \cite{genap, at3d, cvpr23_opt_attack}, attribute editing \cite{semantic_adv, adv_attribute}, and adversarial example generation through makeup transfer \cite{adv_makeup, amt_gan, clip2protect, diff_am}. Both restricted and unrestricted adversarial attacks on FR models have substantially advanced the capabilities of these attacks. However, existing methods for generating adversarial attacks on FR often rely on surrogate models with fixed parameters, which limits the transferability of the crafted adversarial examples. In contrast, our proposed attack method addresses this limitation by diversifying the surrogate model parameters, using models with varying parameters across different epochs.

\section{Methodology}
\subsection{Problem Formulation and Framework}
\noindent \textbf{Problem Formulation.} Let $\gF^{vct}(\x') \in \R ^{r}$ denote the FR model employed by the victim to extract the embedding from a face image $\x'$. We denote $\xs$ and $\xt$ as the source and target images, respectively. The objective of the adversarial attacks explored in our research is to manipulate $\gF^{vct}(\x)$ to misclassify the adversarial example $\xadv$ as the target image $\xt$, while ensuring that $\xadv$ bears a close visual resemblance to $\xs$.
The objective of adversarial attack as delineated in this paper is as follows:
\begin{equation}
	\begin{gathered}		\xadv=\mathop{\arg\min}\limits_{\xadv}\left(\gD\left(\gF^{vct}\left(\xadv\right), \gF^{vct}\left(\bfx^t\right)\right)\right) \quad \\ \text{s.t.} \Vert \xadv - \bfx^s\Vert_p \leq \epsilon
		\label{eq:opt_obj_of_fr}
	\end{gathered}
\end{equation}
where the symbol $\gD$ denotes a predefined distance metric employed for the optimization of adversarial face examples.

In practice, the attacker typically cannot access the model owned by the victim. As a result, black-box attacks play a crucial role in adversarial attacks \cite{wang2021delving, wang2022dst, qian2024dynamic}. A widely adopted approach for black-box attacks involves using a surrogate model $\gF$ to generate adversarial examples and then transfer the crafted adversarial examples to the victim model \cite{vt, bpfa}.
Consequently, the transferability is important and constitutes the key problem studied in this research.

\begin{figure*}[htbp]
	\begin{center}
		\centerline{\includegraphics[width=170mm]{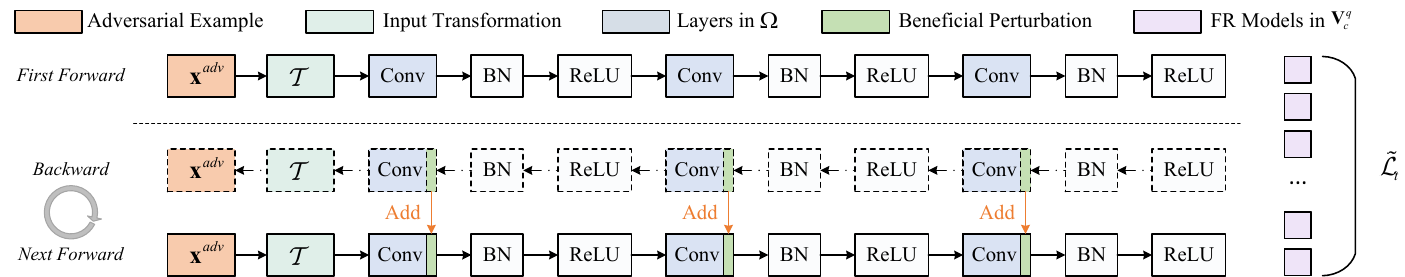}}
		\caption{The framework of the Hard Model Aggregation (HMA). After acquiring a surrogate model set with diverse parameters (i.e., $\bfV^q_c$), we introduce beneficial perturbations with the optimization direction opposite to that of adversarial perturbations onto the feature maps of these diversified surrogate models, transforming them into hard models and aggregate the hard models to increase the transferability.
		}
		\label{fig:hma}
	\end{center}
    \vspace{-0.8cm}
\end{figure*}

\noindent \textbf{Framework Overview.} 
To enhance the transferability of adversarial examples, we propose a novel method called Diverse Parameters Augmentation (DPA). DPA augments surrogate models by incorporating both pre-trained and randomly initialized parameters, resulting in a more diverse and expansive set of surrogate models. Specifically, DPA consists of two key stages: Diverse Parameters Optimization (DPO) and Hard Model Aggregation (HMA). 
In the DPO stage, we diversify the surrogate model parameters by combining pre-trained and random initializations, as illustrated in \cref{fig:dpo}. In the HMA stage, we apply beneficial perturbations to the feature maps of the diversified surrogate models and combine them to achieve a higher degree of augmentation, as shown in \cref{fig:hma}. In the following, we will introduce our proposed DPA attack method in detail.

\subsection{Diverse Parameters Optimization}\label{sec:dpo}
The parameter diversity of surrogate models is crucial for the transferability of crafted adversarial examples. Previous parameter-based augmentation adversarial attack methods solely initialize parameters with pre-trained values, thereby limiting the parameter diversity of surrogate models \cite{lgv, sasd}. In contrast, the key innovation of the stage of our proposed attack method is the diversification of initialized parameters using both pre-trained and randomly initialized parameters. Following initialization, we augment the parameters of the surrogate models using the parameters in the intermediate training process to diversify the surrogate models, thereby improving the transferability of the adversarial examples.

Let $\x$ be a batch of face images, $\w \in \R^{s \times r}$ be the parameters of the FR head, $b$ be the batch size of $\x$, and $s$ be the class number in the training dataset. To calculate the loss function, we should first calculate the cosine similarity matrix $\bfr  \in \R^{b \times s}$ between $\gF \left(\x\right) \in \R^{b \times r}$ and the parameters of the FR head $\w^{\top}$ using the following formula:
\begin{equation}
	\begin{gathered}
        \bfr = \cos \bfa = \widebar{\e} \widebar{\w}^{\top}
		\label{eq:m}
	\end{gathered}
\end{equation}
where $\widebar{\e}=\frac{\gF \left(\x\right)}{\left \Vert \gF \left(\x\right) \right \Vert}$, $\widebar{\w}=\frac{\w}{\left \Vert \w \right \Vert}$ and $\bfa$ is the angle between $\gF \left(\x\right)$ and $\w^{\top}$.
After obtaining the cosine similarity matrix by \cref{eq:m}, we can get the sine similarity between $\gF \left(\x\right)$ and $\w$ using the following equation:
\begin{equation}
	\begin{gathered}
        \sin \bfa = \sqrt{1.0 - \cos^ 2 \bfa }
		\label{eq:sin_fx_weight}
	\end{gathered}
\end{equation}

Using the cosine and sine similarity matrix, the formula to get the additive angular margin cosine similarity matrix can be expressed as:
\begin{equation}
	\begin{gathered}
        \cos \left( \bfa + m\right) = \cos \bfa \cos m - \sin \bfa \sin m
		\label{eq:cos_alpha_m}
	\end{gathered}
\end{equation}
where $m$ is the margin value for increasing the discrimination of the FR model \cite{arcface}.
Let $\bfy \in \R^{b}$ be the labels of the current batch. By utilizing $\bfy$, we can calculate its one-hot encoded matrix using the following formula:
\begin{equation}
	\begin{gathered}
        \widehat{\bfh} = \Psi \left(\bfy\right) \in \R^{b \times s}
		\label{eq:widehat_h}
	\end{gathered}
\end{equation}
where $\Psi \left( \cdot \right)$ is the one-hot encode operation.
Next, we use the following function to calculate the output of the head:
\begin{equation}
	\begin{gathered}
        \bfq = d\left(\widehat{\bfh} \odot \bfp + \left(1 - \widehat{\bfh}\right) \odot \bfr\right) \\
        \bfp = 
        \left\{
        \begin{array}{ll}
        \cos \left( \bfa + m\right) & \text{s.t.} \quad \bfa < \pi - m, \\
        \cos \bfa - m \sin m & \text{s.t.} \quad \bfa \ge \pi - m,
        \end{array}
        \right.
		\label{eq:bar_h}
	\end{gathered}
\end{equation}
where $d$ is a pre-defined scale factor.
Using the output of the head, the formula to calculate the loss function can be expressed as the following:
\begin{equation}
	\begin{gathered}
        \gL \left(\x, \bfy\right) = - \frac{1}{b}\sum_{i=1}^{b} \log \frac{e^{\bfq_{i, \bfy_{i}}}}{\sum_{j=1}^{s} e^{\bfq_{i,j}}}
		\label{eq:l_t}
	\end{gathered}
\end{equation}
Let $c$ and $\bfv$ be the number of the training epoch and the parameters of the FR backbone, respectively. Utilizing \cref{eq:l_t}, we can get the parameters set \cite{arcface}:
\begin{equation}
	\begin{gathered}
        \bfv_i, \bfw_i = \varphi \left(\argmin_{\bfv, \bfw} \gL \left(\x, \bfy\right),i\right) \quad i \in \left\{1, 2, ..., c\right\} \\
        \bfV_c = \left\{\bfv_1, \bfv_2, ..., \bfv_{c-1}, \bfv_{c}\right\} \quad \bfW_c = \left\{\bfw_1, \bfw_2, ..., \bfw_{c-1}, \bfw_{c}\right\}
		\label{eq:parameters_set}
	\end{gathered}
\end{equation}
where $\varphi \left(\cdot, i\right)$ is the operation that obtain the $\bfv$ and $\bfw$ parameters in the $i$th epoch. We denote $\bfv_0$ and $\bfw_0$ as the initialized parameters of the FR model and head, respectively. If $\bfv_0$ and $\bfw_0$ are fixed, $\bfV_c$ and $\bfW_c$ become deterministic aside from minor random factors in the training process of \cref{eq:parameters_set}. Therefore, there exists a mapping between $\left\{\bfV_c, \bfW_c \right\}$ and $\left\{\bfv_0, \bfw_0\right\}$ that can be expressed as:
\begin{equation}
	\begin{gathered}
        \left\{\bfV_c, \bfW_c \right\} = \digamma \left(\left\{\bfv_0, \bfw_0\right\}, i\right)
		\label{eq:digamma}
	\end{gathered}
\end{equation}

Existing adversarial attack methods based on parameter augmentation typically initialize the parameters $\left\{\bfv_0, \bfw_0\right\}$ solely with pre-trained values \cite{lgv, sasd}. This constraint diminishes the diversity within the augmented surrogate model set.
In contrast, we opt to initialize $\left\{\bfv_0, \bfw_0\right\}$ with diverse parameters to enhance the diversity of the augmented surrogate model set, thereby improving transferability.
The $\left\{\bfV_c, \bfW_c \right\}$ of our proposed attack can be expressed as:
\begin{equation}
	\begin{gathered}
        \left\{\bfV_c^p, \bfW_c^p \right\} = \digamma \left(\left\{\bfv_0^p, \bfw_0^p\right\}, i\right) \\
        \left\{\bfV_c^a, \bfW_c^a \right\} = \digamma \left(\left\{\bfv_0^a, \bfw_0^a\right\}, i\right)
		\label{eq:our_VW}
	\end{gathered}
\end{equation}
where $\bfv_0^p$ represents the pre-trained parameters, while $\bfw_0^p$, $\bfv_0^a$ and $\bfw_0^a$ are randomly initialized. Using \cref{eq:our_VW}, we can obtain the more diverse parameters for the FR model and head. After obtaining $\bfV_c^p$ and $\bfV_c^a$, we use the following set of the parameters to craft the adversarial examples:
\begin{equation}
	\begin{gathered}
        \bfV_c^q = \left\{\bfv_0^p, \bfv_1^p, \bfv_{j_1}^p, ..., \bfv_{c}^p\right\} \bigcup \left\{\bfv_1^a, \bfv_{j_1}^a, \bfv_{j_2}^a, ..., \bfv_{c}^a\right\}\\
        {\rm s.t.} j \in \left\{1, 2, ..., c \vert j \bmod \kappa = 1\right\}
		\label{eq:wcs}
	\end{gathered}
\end{equation}
where $\kappa = \lfloor \sqrt{c} \rfloor$ determines the epoch interval to select the parameters. 
The overall framework and pseudo-code of the DPO stage in our proposed method is illustrated in \cref{fig:dpo} and \cref{algo:dpo}, respectively.

\begin{algorithm}
	\renewcommand{\algorithmicrequire}{\textbf{Input:}}
	\renewcommand{\algorithmicensure}{\textbf{Output:}}
	\caption{Diverse Parameters Optimization (DPO)} 
	\label{algo:dpo} 
	\begin{algorithmic}[1]
        \REQUIRE Diverse initial parameters $\bfv_0^p$, $\bfw_0^p$, $\bfv_0^a$, and $\bfw_0^a$, the number of epochs \( c \), the optimizer \( \bfT \), the FR training dataset \( \zeta \), loss function $\gL$.
		\ENSURE Augmented parameter set $\bfV_c^q$.
		\STATE $\bfP=\left\{\bfv_0^p, \bfw_0^p\right\}$, $\bfA=\left\{\bfv_0^a, \bfw_0^a\right\}$, $\bfV_c^q=\left\{\bfv_0^p\right\}$
		\FOR{$\bfJ \in \left\{\bfP, \bfA\right\}$}
		\STATE $\left\{\bfv,\bfw\right\}=\bfJ$
		\STATE $\gF = \Psi \left(\bfv\right)$ \COMMENT{Map the parameters to the models.}
        \FOR{$i=1,..., c$}
        \FOR{$\bfx \in \zeta$}
        \STATE Calculate $\gL$ using \cref{eq:l_t}.
        \STATE Backward($\bfT$, $\gL$)
        \COMMENT{Backpropagation using $\bfT$.}
        \STATE Update the parameters $\bfv,\bfw$.
        \ENDFOR
        \IF{the \(\bfv\) satisfies the condition outlined in \cref{eq:wcs}}
        \STATE Add($\bfV_c^q$, $\bfv$)
        \COMMENT{Incorporate \(\bfv\) into \(\bfV_c^q\).}
        \ENDIF
        \ENDFOR
		\ENDFOR
	\end{algorithmic} 
\end{algorithm}
\subsection{Hard Model Aggregation}\label{sec:hma}
Hard models \cite{bpfa} refer to models with higher loss values compared to normal models, designed to simulate the effect of hard samples, which are hard to the optimization process of the adversarial examples. Since hard samples can improve the generalization of models, we leverage multiple hard models to enhance the transferability of the surrogate models with diverse parameters by adding beneficial perturbations \cite{Beneficial_Perturbation_Network} onto the pre-defined feature maps of the surrogate models.

In the following, we will provide a detailed introduction to the HMA stage.
Let $\Psi$ be the mapping from the parameters to the corresponding models that can be expressed as:
\begin{equation}
	\begin{gathered}
        \bfF = \Psi \left(\bfV_c^q\right) = \left\{\gF_1, \gF_2, ..., \gF_{g}\right\}
		\label{eq:bfF}
	\end{gathered}
\end{equation}
where $g=\left \vert \bfV_c^q \right \vert$.
Hard sample augmentation has demonstrated significant effectiveness in enhancing model generalization \cite{teach_aug, adv_aug}. Based on the relationship between adversarial attack tasks and training tasks \cite{mim, ni_fgsm}, using hard models for augmentation can lead to improved transferability \cite{bpfa}. Consequently, by transforming vanilla parameter-augmented surrogate models into hard parameter-augmented surrogate models, we can achieve more transferable results. To accomplish this transformation, we use the following formula to calculate the loss function:
\begin{equation}
	\begin{gathered}
        \widetilde{\gL}_t = \frac{1}{g}\sum_{i=0}^g \left \Vert \phi \left(\gH_i \left(\gT\left(\xadv_t\right)\right)\right) - \phi \left(\gF_i \left(\xt\right)\right)\right \Vert \\
        {\rm s.t.} \quad \gF_i \in \bfF \quad t \in \left\{1,2, ..., n\right\}
		\label{eq:lc}
	\end{gathered}
\end{equation}
where $\xadv_t$ denotes the adversarial example generated during the $t$th iteration, the variable $n$ represents the maximum number of iterations allocated for crafting the adversarial examples, $\phi \left(\cdot\right)$ denotes the normalization operation, $\gT$ signifies the input transformation, and $\gH$ is the corresponding hard model \cite{bpfa} of $\gF$, which can be expressed as:
\begin{equation}
	\begin{gathered}
        \gH \left(\xadv_t\right) = 
        \left\{
        \begin{array}{ll}
        \gF \left(\xadv_t\right) \quad {\rm s.t.} \quad t=1, \\
        \chi \left(\gF \left(\xadv_t\right), \Omega\right)\quad {\rm s.t.} \quad t > 1,
        \end{array}
        \right.
		\label{eq:gH}
	\end{gathered} 
\end{equation}
where $\chi$ is the mapping for adding beneficial perturbations on the feature maps in the forward propagation whose formula can be expressed as:
\begin{equation}
	\begin{gathered}
         \chi \left(\gF \left(\xadv_t\right), \Omega\right) : \omega = \omega + \eta {\rm sign} \left(\nabla_\omega\widetilde{\gL}_{t-1}\right)\\
         {\rm s.t.} \quad \omega \in \Omega
		\label{eq:chi}
	\end{gathered}
\end{equation}
where $\Omega$ is the pre-defined set of layers for adding beneficial perturbations.
Let $\epsilon$ be the maximum allowable perturbation.
Utilizing $\widetilde{\gL}_t$, we can craft the adversarial face examples using following formula:
\begin{equation}
	\begin{gathered}
        \xadv_{t+1} = \prod_{\xs, \epsilon}\left(\xadv_{t}-\beta {\rm sign} \left(\nabla_{\xadv_{t}}\widetilde{\gL}_{t}\right)\right)
		\label{eq:opt_ax}
	\end{gathered}
\end{equation}
where $\beta$ denotes the step size used to construct the adversarial examples, while $\prod$ represents the clipping operation that confines the pixel values of the generated adversarial examples within the range $\left[\xs - \epsilon, \xs + \epsilon\right]$.
The framework of the HMA is illustrated in \cref{fig:hma}.

In the following, we present the pseudo-code for the HMA. For the sake of clarity, we focus on neural networks with a single branch in their computational graphs. In the case of neural networks with multiple branches, the HMA algorithm remains largely unchanged, except for the incorporation of mechanisms to handle multiple branches. Before proceeding, we introduce some necessary definitions. Let $\Phi$ denote the index set of pre-selected layers to which beneficial perturbations will be added:
\begin{equation}
	\Phi=\varkappa \left(\Omega\right)
\end{equation}
where \(\varkappa\) represents the mapping from the pre-defined set of layers for adding beneficial perturbations \(\Omega\), to their corresponding layer index set $\Phi$. Let \(f^i\) denote the \(i\)th layer within the model \(\gF\), and let $z$ represent the total number of layers in \(\gF\). We define the segment of \(\gF\) from layer \(f^i\) to layer \(f^j\) as follows:

\begin{equation}
	\gF^{i,j}=f^i \circ f^{i+1} \circ ... \circ f^{j-1} \circ f^j  
\end{equation}

The pseudo-code for the HMA stage of our proposed method is presented in \cref{algo:hma}.

\begin{algorithm}
	\renewcommand{\algorithmicrequire}{\textbf{Input:}}
	\renewcommand{\algorithmicensure}{\textbf{Output:}}
	\caption{Hard Model Aggregation (HMA)} 
	\label{algo:hma} 
	\begin{algorithmic}[1]
		\REQUIRE The source image $\xs$, the target image $\xt$, the mapping from the parameters to the corresponding models $\Psi$, the maximum number of iterations $n$, the index set of pre-selected layers to be added beneficial perturbations $\Phi$, the step size of the beneficial perturbations $\eta$, the step size of the adversarial perturbations $\beta$, the maximum permissible magnitude of perturbation $\epsilon$, the parameter set $\bfV_c^q$, the total number of layers in a single surrogate model $z$, normalization operation $\phi$. 
		\ENSURE An adversarial example $\xadv_{n}$
		\STATE $\xadv_0=\xs$, $u=\vert\Phi\vert$, $s_1=1$, $g=\vert\bfV_c^q\vert$
        \STATE $\bfF = \Psi \left(\bfV_c^q\right)$
        \COMMENT{Acquire the diversified models set.}
		\FOR{$t=1 ,..., n$}
        \FOR{$i=1 ,..., g$}
		\STATE $\gF=\bfF_i$
        \COMMENT{Derive the parameter-augmented model.}
        \STATE $\omega^{t,0}=\gT \left(\xadv_{t-1}\right)$
		\FOR{$j=1 ,..., u$}
		\STATE $s_2=\Phi_j$
		\STATE $\omega^{t,j}=\gF^{s_1,s_2}\left(\omega^{t,j-1}\right)$
		\STATE $s_1=s_2$
		\IF{$t \neq 1$}
		\STATE $\omega^{t,j}=\omega^{t,j}+\eta {\rm sign}\left(\nabla_{\omega^{t-1,j}}\widetilde{\gL}_{t-1}\right)$
		\ENDIF
        \ENDFOR
		\STATE $\widetilde{\gL}^i=\Vert\phi\left(\gF^{\Phi_u, z}\left(\omega^{t,u}\right)\right)-\phi\left(\gF\left(\xt\right)\right)\Vert^2_2$
		\ENDFOR
        \STATE $\widetilde{\gL}_t=\frac{1}{g}\sum_{i=1}^g \widetilde{\gL}^i$
        \COMMENT{Compute the loss function utilizing the parameter-augmented models.}
		\STATE $\xadv_{t} = \prod \limits_{\xs, \epsilon}\left(\xadv_{t-1}-\beta {\rm sign}\left(\nabla_{\xadv_{t-1}}\widetilde{\gL}_t\right)\right)$ 
		\ENDFOR
	\end{algorithmic} 
\end{algorithm}

\section{Experiments}
\subsection{Experimental Setting}\label{experimental_setting}
\textbf{Datasets.} We opt to use the LFW \cite{lfw}, and CelebA-HQ \cite{celeba_hq} for our experiments to verify the effectiveness of our proposed attack method. LFW serves as an unconstrained face dataset for FR. CelebA-HQ consists of face images with high visual quality. The LFW and CelebA-HQ utilized in our experiments are identical to those employed in \cite{bpfa,adv_restore,adv_pruning}. For the parameter augmentation process, we select BUPT-Balancedface~\cite{rfw,wang2021meta,wang2020mitigating,wang2018deep} as our training dataset.

\noindent \textbf{Face Recognition Models.}
The normal trained FR models employed in our experiments encompass IR152~\cite{resnet}, IRSE50~\cite{irse50}, FaceNet~\cite{facenet}, MobileFace~\cite{arcface}, CurricularFace~\cite{curricular_face}, MagFace~\cite{mag_face}, ArcFace~\cite{arcface}, CircleLoss~\cite{circle_loss}, MV-Softmax~\cite{mv_softmax}, and NPCFace~\cite{npcface}. Specifically, IR152, FaceNet, IRSE50, and MobileFace are the same models utilized in~\cite{adv_makeup, amt_gan, bpfa, adv_restore}. CurricularFace, MagFace, ArcFace, CircleLoss, MV-Softmax, and NPCFace are the official models provided by FaceX-ZOO~\cite{facex_zoo}. Furthermore, we integrate adversarial robust FR models into our experiments, denoted as IR152$^{adv}$, IRSE50$^{adv}$, FaceNet$^{adv}$, and MobileFace$^{adv}$, which correspond to the models used in~\cite{bpfa}.

\noindent \textbf{Attack Setting.}
We set the maximum allowable perturbation magnitude $\epsilon$ to 10, based on the $L_{\infty}$ norm bound, without any specific emphasis. In addition, we specify the maximum number of iterative steps as 200.
More detailed attack settings are provided \textit{in the supplementary}.

\noindent \textbf{Evaluation Metrics.}
We utilize Attack Success Rate (ASR) as the metric to assess the efficacy of various adversarial attacks. The ASR represents the ratio of adversarial examples that successfully evade the victim model to the total number of adversarial examples generated.
In calculating the ASR, we determine the threshold based on a FAR@0.001 on the entire LFW dataset for each victim model.

\noindent \textbf{Compared methods.}
Our proposed attack is a form of restricted adversarial attack designed to expose vulnerabilities within FR systems. It would be inequitable to juxtapose our method with unrestricted attacks, which do not impose limits on the magnitude of adversarial perturbations. Consequently, we opt to benchmark our approach against other restricted attacks on FR systems that are explicitly malicious at attacking the systems \cite{dfanet} \cite{bpfa}, as well as state-of-the-art transfer attacks \cite{dim} \cite{ssa6} \cite{lgv} \cite{sia} \cite{bsr}.

\subsection{Comparison Studies}
\begin{figure}[htbp]
	\begin{center}
		\centerline{\includegraphics[width=\columnwidth]{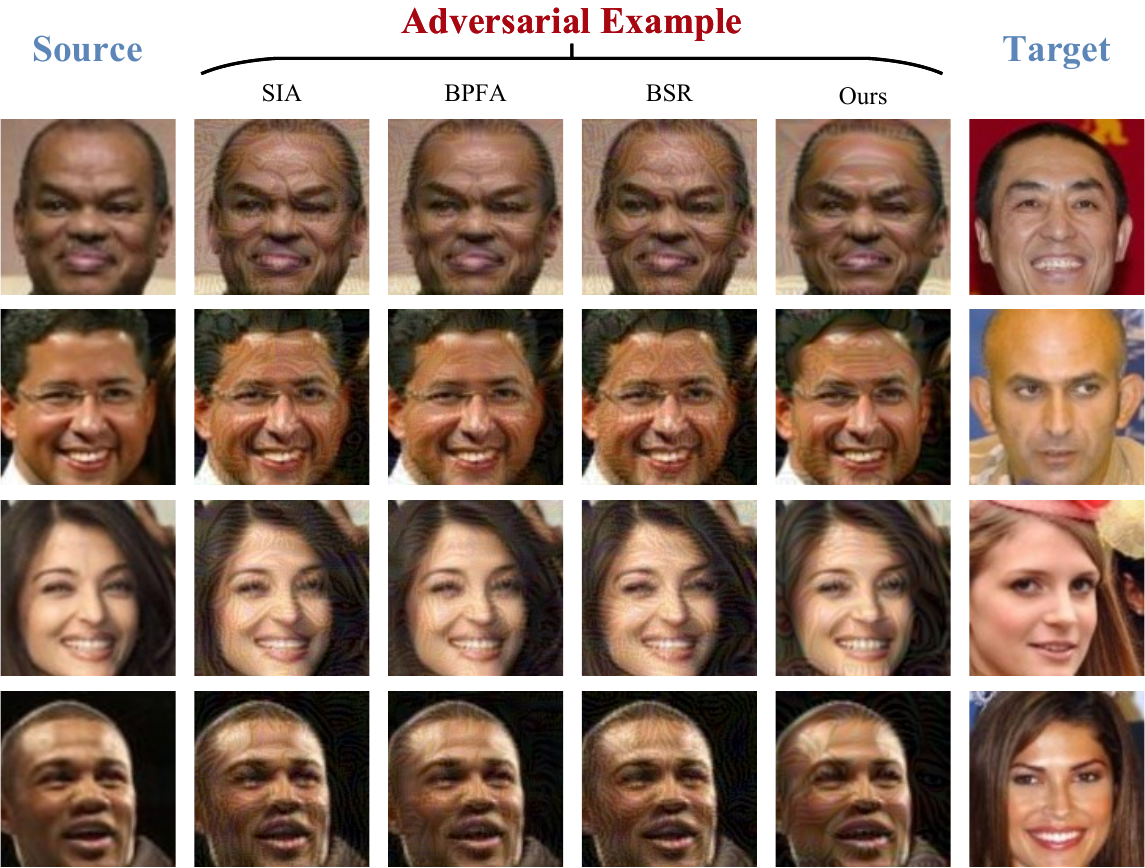}}
		\caption{The illustration of adversarial examples crafted by various attacks. First column: some  source images. Last column: the corresponding target images.
        The second to fifth columns exhibit the corresponding adversarial face examples crafted by SIA \cite{sia}, BPFA \cite{bpfa}, BSR \cite{bsr}, and Our proposed attack, respectively.
		}
		\label{fig:ax_show}
	\end{center}
    \vspace{-0.8cm}
\end{figure}

\textbf{DPA achieves the best black-box attack results on both normally trained and adversarial robust models.}
To verify the effectiveness of our proposed attack method, we craft the adversarial examples on the LFW and CelebA-HQ dataset. The black-box performance with MobileFace as the surrogate models on the LFW dataset are demonstrated in \cref{tab:compare_lfw_m}. Some of the adversarial examples are demonstrated in \cref{fig:ax_show}.
More experimental results on the LFW dataset, as well as those on the CelebA-HQ dataset, are provided \textit{in the supplementary}.
These results demonstrate that our proposing method consistently outperforms the baseline attack, thereby highlighting its effectiveness in improving the transferability of adversarial face examples.

\begin{table}[]
\caption{Comparisons of black-box ASR (\%) results for attacks using MobileFace as the surrogate model on the LFW dataset. I, S, F, M denote IR152, IRSE50, FaceNet, and MobileFace, respectively.}
\label{tab:compare_lfw_m}
\centering
\adjustbox{width=\columnwidth}{
\begin{tabular}{l|ccccccc}
\hline
Attacks & I         & S        & F       & I$^{adv}$    & S$^{adv}$     & F$^{adv}$  & M$^{adv}$         \\ \hline
FIM~\cite{fim}     & 5.3           & 73.4          & 7.5           & 2.5           & 4.5           & 2.8           & 10.9          \\
DI~\cite{dim}      & 18.4          & 97.3          & 32.9          & 10.6          & 18.2          & 10.2          & 38.6          \\
DFANet~\cite{dfanet}  & 7.0           & 86.4          & 11.9          & 3.6           & 6.4           & 3.9           & 16.8          \\
VMI~\cite{vt}     & 13.6          & 96.0          & 20.2          & 7.6           & 12.8          & 7.6           & 32.3          \\
SSA~\cite{ssa6}     & 13.8          & 96.4          & 19.6          & 5.5           & 13.1          & 7.2           & 31.5          \\
SIA~\cite{sia}     & 15.7          & 96.8          & 26.7          & 8.1           & 14.5          & 9.1           & 35.3          \\
BPFA~\cite{bpfa}    & 15.8          & 95.5          & 17.8          & 5.9           & 11.0          & 5.1           & 28.4          \\
BSR~\cite{bsr}     & 5.4           & 74.2          & 9.5           & 2.9           & 5.9           & 4.9           & 11.4          \\
Ours    & \textbf{67.7} & \textbf{98.2} & \textbf{90.8} & \textbf{55.4} & \textbf{66.4} & \textbf{42.6} & \textbf{71.6} \\ \hline
\end{tabular}
}
\end{table}

\begin{figure}[t]
	\begin{center}
		\centerline{\includegraphics[width=80mm]{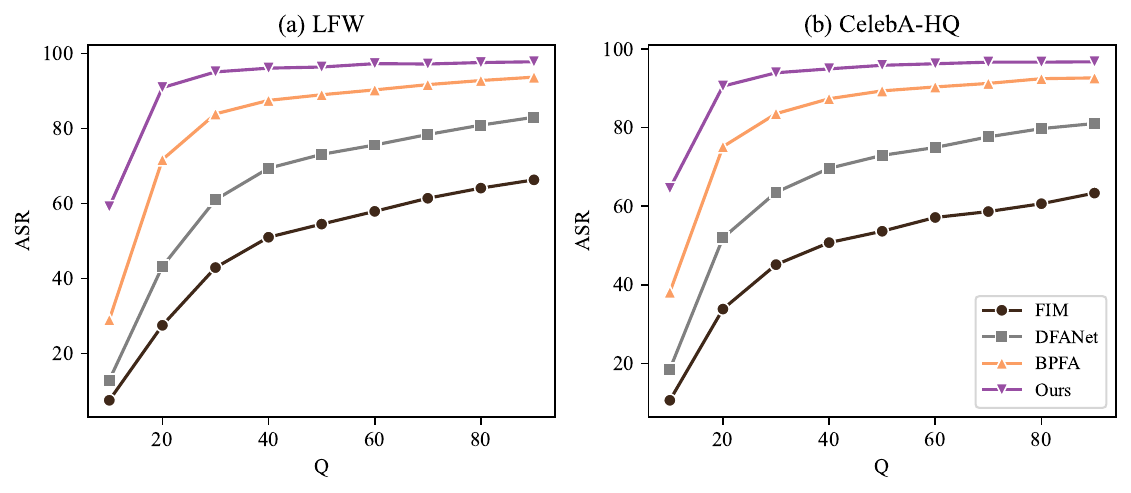}}
		\caption{Performance of ASR across various JPEG Q values: (a) Results on the LFW dataset. (b) Results on the CelebA-HQ dataset.
		}
		\label{fig:jpeg_compress_study}
	\end{center}
    \vspace{-0.8cm}
\end{figure}

\noindent \textbf{DPA achieves superior black-box performance under JPEG compression.}
JPEG compression is a widely adopted method for image compression during transmission, and it also serves as a defense mechanism against adversarial examples. To evaluate the effectiveness of our proposed attack under JPEG compression, we employ MobileFace as the surrogate model and IRSE50 as the victim model. We assess the attack performance on the LFW and CelebA-HQ datasets, with the results presented in \cref{fig:jpeg_compress_study}. These results indicate that our proposed attack method consistently outperforms the baseline attack methods across varying levels of JPEG compression, thereby underscoring the robustness of our proposed attack method under such conditions.

\noindent \textbf{DPA demonstrates better black-box performance compared to the parameter-based augmented adversarial attack.}
The LGV method is an effective parameter-based augmented adversarial attack technique designed to enhance transferability. Our proposed attack incorporates parameter augmentation, making LGV an appropriate baseline. We selected MobileFace as the surrogate model and generated adversarial examples on the LFW dataset. The results are presented in \cref{tab:compare_w_param_aug_method}. \cref{tab:compare_w_param_aug_method} clearly shows that our proposed attack outperforms the baseline, further validating the effectiveness of our method.
The primary limitation of LGV stems from its random selection of one model per iteration, which introduces instability and adversely impacts black-box ASR \cite{mim, vt}. Furthermore, LGV suffers from a lack of diverse initialization parameters. These shortcomings significantly impair the overall ASR performance of LGV.

\begin{table}[]
\caption{Comparison of black-box ASR (\%) results using the parameter-based augmented adversarial attack method as the baseline on the LFW dataset. ASR$^{adv}$ represents the average attack success rate on adversarial robust models.
}
\label{tab:compare_w_param_aug_method}
\centering
\begin{tabular}{c|cccc}
\hline
         & IR152         & IRSE50        & FaceNet       & ASR$^{adv}$ \\ \hline
Baseline & 4.8           & 19.6          & 15.7          & 7.2         \\
Ours     & \textbf{67.7} & \textbf{98.2} & \textbf{90.8} & \textbf{59.0} \\ \hline
\end{tabular}
\end{table}

\begin{figure}[t]
	\begin{center}
		\centerline{\includegraphics[width=80mm]{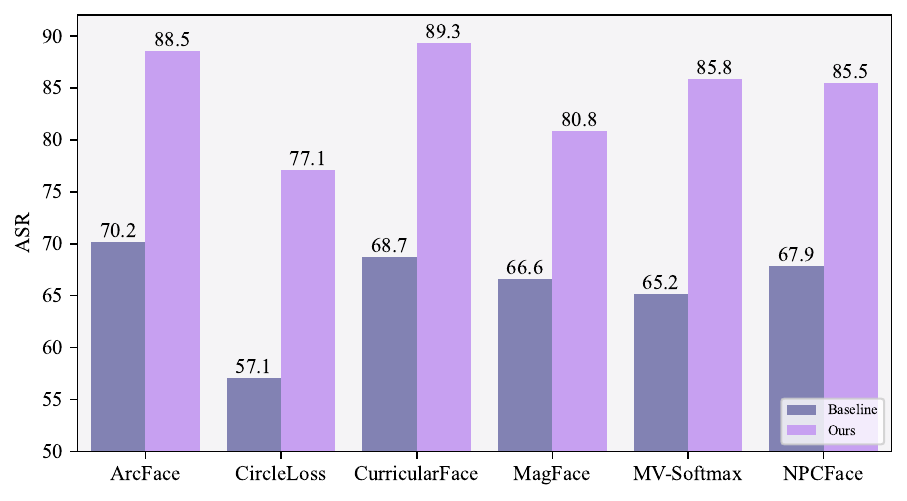}}
		\caption{ASR on victim models trained with various algorithms, using FaceNet as the surrogate model on the LFW dataset.
		}
		\label{fig:compare_study_more_fr_models}
	\end{center}
    \vspace{-0.8cm}
\end{figure}

\noindent \textbf{DPA achieves superior black-box performance on FR models trained with various algorithms.}
In order to substantiate the effectiveness of our proposed attack method across various FR models, we conducted additional experiments using BSR \cite{bsr} as the Baseline and FaceNet as the surrogate model. These experiments adhered to the same settings detailed in \cref{tab:compare_lfw_f} \textit{in the supplementary}. The ASR across multiple FR models is illustrated in \cref{fig:compare_study_more_fr_models}. As shown in \cref{fig:compare_study_more_fr_models}, the ASR of the adversarial examples generated by our proposed method exceeds that of the Baseline, thereby reinforcing the efficacy of our proposed method.

\subsection{Ablation Studies}

\noindent \textbf{The ablation studies on each stage of our proposed attack method.}
Our proposed attack method consists of two stages. To verify the effectiveness of each stage, we conduct ablation studies on the stages. Initially, we craft adversarial examples using only the surrogate model, which we denote as `Vanilla'. Next, we generate adversarial examples using the ensemble of surrogate models obtained in the DPO stage (\ie $\bfF$), denoted as DPO. After incorporating the HMA stage, the complete attack method is denoted as `DPO + HMA'. The results are presented in \cref{tab:ablation_study}. \cref{tab:ablation_study} demonstrates that the addition of the DPO stage results in an increase in black-box ASR, showcasing the effectiveness of the DPO stage in enhancing transferability. Further incorporation of the HMA stage leads to an additional improvement in attack performance, underscoring the effectiveness of the HMA stage in boosting black-box performance. These results collectively demonstrate the effectiveness of each stage in our proposed attack method in improving the transferability.

\begin{table}[]
\caption{Ablation study of black-box ASR (\%) results on the LFW dataset using MobileFace as the surrogate model. ASR$^{adv}$ denotes the average attack success rate on adversarial robust models.
}
\label{tab:ablation_study}
\centering
\begin{tabular}{c|cccc}
\hline
Attacks   & IR152         & IRSE50        & FaceNet       & ASR$^{adv}$   \\ \hline
Vanilla   & 5.3           & 73.4          & 7.5           & 5.2           \\
DPO       & 35.9          & 91.3          & 67.9          & 32.9          \\
DPO + HMA & \textbf{67.7} & \textbf{98.2} & \textbf{90.8} & \textbf{59.0} \\ \hline
\end{tabular}
\end{table}

\noindent \textbf{The ablation studies on the effectiveness of multiple surrogate models at different epochs in improving transferability.}
Since our proposed attack method utilizes models from intermediate training epochs to craft adversarial examples, it is essential to verify the effectiveness of this approach compared to using only models from the final training epoch. We conduct ablation experiments on these two approaches using the LFW dataset and MobileFace as the surrogate model. We employ the DPO process to obtain the parameter sets $\bfV_c^q$ and $\bfV_c^m = \left\{\bfv_0^p, \bfv_{c}^p, \bfv_{c}^a\right\}$. Next, we map $\bfV_c^m$ into its corresponding model set $\bfF^m = \Psi \left(\bfV_c^m\right)$. We then craft adversarial examples by ensembling the models from $\bfV_c^q$ (\ie $\bfF$) and $\bfF^m$, respectively. The results are presented in \cref{tab:inter_and_final_epoch}. \cref{tab:inter_and_final_epoch} shows that the performance of the ensemble of $\bfF$ surpasses that of the ensemble of $\bfF^m$, demonstrating the effectiveness of using models from intermediate training epochs to enhance the transferability.

\begin{table}[]
\caption{Ablation study of black-box ASR (\%) results to verify the effectiveness of multiple surrogate models at different epochs in improving transferability. ASR$^{adv}$ denotes the average attack success rate on adversarial robust models.
}
\label{tab:inter_and_final_epoch}
\centering
\begin{tabular}{c|cccc}
\hline
Attacks       & IR152         & IRSE50        & FaceNet       & ASR$^{adv}$   \\ \hline
$\bfF^m$      & 20.3          & 86.7          & 33.6          & 17.8          \\
$\bfF$ (Ours) & \textbf{35.9} & \textbf{91.3} & \textbf{67.9} & \textbf{32.9} \\ \hline
\end{tabular}
\end{table}

The hyper-parameter analysis studies of our proposed method are \textit{in the supplementary}.
\section{Conclusion}
We present an innovative advancement in the field of adversarial attacks on FR through the introduction of the Diverse Parameters Augmentation (DPA) attack method. By addressing the problems of traditional adversarial attacks, particularly the lack of diverse parameters and the exclusion of the FR head, DPA enhances the transferability of adversarial face examples. In addition, DPA emphasizes the importance of integrating both pre-trained and randomly initialized parameters, providing a fresh perspective on augmentation-based adversarial attacks.
The extensive experimental results demonstrate the effectiveness of our proposed DPA attack across multiple datasets and target models. We anticipate that our proposed method will illuminate potential directions for advancing research in adversarial attacks on FR.
\vspace{-0.1cm}
\section{Acknowledgment}
This work was supported in part by the Natural Science Foundation of China under Grant 62372203 and 62302186, in part by the Major Scientific and Technological Project of Shenzhen (202316021), in part by the National key research and development program of China (2022YFB2601802), in part by the Major Scientific and Technological Project of Hubei Province (2022BAA046, 2022BAA042).
{
    \small
    \bibliographystyle{ieeenat_fullname}
    \bibliography{main}
}
\clearpage
\setcounter{page}{1}
\maketitlesupplementary
\section{Appendix}
\noindent \textbf{Overview.} The supplementary includes the following sections:
\begin{itemize}
    \item \textbf{\cref{sec:compute_asr}.} Computation Methodology for Attack Success Rate.
    \item \textbf{\cref{sec:more_detailed_attack_setting}.} More Detailed Attack Settings.
    \item \textbf{\cref{sec:more_exp_on_lfw}.} More Comparison Studies on LFW.
    \item \textbf{\cref{sec:celeba_hq}.} Comparison Studies on CelebA-HQ.
    \item \textbf{\cref{sec:hyper_sen_study}.} Hyper-parameter Analysis Studies.
    \item \textbf{\cref{sec:visual_quality}.} Visual Quality Study.
    \item \textbf{\cref{sec:broader_impact}.} Ethics and Potential Broader Impact.
\end{itemize}

\subsection{Computation Methodology for Attack Success Rate} \label{sec:compute_asr}
In our study, the Attack Success Rate (ASR) is determined using the following formula:
\begin{equation}
	{\rm ASR}=\frac{\sum_{i=1}^{N_p}\mathds{1}\left(\widetilde{\gD}\left(\gF^{vct}\left(x^{adv}\right), \gF^{vct}\left(x^{t}\right)\right)<t^i\right)}{N_p}\label{eq:asr_impersonation}
\end{equation}
where the notation $\widetilde{\gD}$ designates a predefined distance metric for assessing the performance of adversarial face examples, $N_p$ denotes the total count of face pairs, and $t^i$ signifies the attack threshold.

\subsection{More Detailed Attack Settings} \label{sec:more_detailed_attack_setting}
In the DPO stage, we set the learning rate to 0.1. In the HMA stage, we specifically target convolutional layers to introduce beneficial perturbations. We maintain the step size for adversarial perturbations $\beta$ at a fixed value of 1. We have set the scale factor \( d \) to 32.0 and the margin \( m \) to 0.5. We employ the SGD optimizer for model augmentation.

For the tables and figures mentioned—\cref{tab:compare_lfw_s}, \cref{tab:compare_lfw_f}, \cref{tab:compare_lfw_m}, \cref{tab:compare_lfw_i}, \cref{tab:compare_celeba_hq_s}, \cref{tab:compare_celeba_hq_f}, \cref{tab:compare_celeba_hq_m}, \cref{tab:compare_celeba_hq_i}, \cref{fig:visual_quality_study}, \cref{fig:jpeg_compress_study}, and \cref{fig:compare_study_more_fr_models}, we determine setting $c$ to 35, which corresponds to the optimal value from the left plot in \cref{fig:m_and_step}, and setting the step size of beneficial perturbations $\eta$ to 8e-4, as indicated by the optimal value from the right plot in \cref{fig:m_and_step}.

Regarding the bottom portion of \cref{fig:motivation}, we have configured the settings for LGV according to the same hyperparameters as specified in \cref{tab:compare_w_param_aug_method}. Similarly, the settings for DI, BPFA, and DPA are aligned with those detailed in \cref{tab:compare_lfw_m}.

\subsection{More Comparison Studies on LFW} \label{sec:more_exp_on_lfw}
To validate the efficacy of our proposed attack method, we generate adversarial examples using IR152, IRSE50, and FaceNet as surrogate models on the LFW dataset. The black-box performance is presented in \cref{tab:compare_lfw_i}, \cref{tab:compare_lfw_s}, and \cref{tab:compare_lfw_f}, respectively. Our method consistently outperforms baseline attacks, demonstrating its effectiveness in improving the transferability of adversarial examples.

\begin{table}[]
\caption{Comparisons of black-box ASR (\%) results for attacks using IR152 as the surrogate model on the LFW dataset. I, S, F, M denote IR152, IRSE50, FaceNet, and MobileFace, respectively.}
\label{tab:compare_lfw_i}
\centering
\adjustbox{width=\columnwidth}{
\begin{tabular}{l|ccccccc}
\hline
Attacks & S        & F       & M            & I$^{adv}$    & S$^{adv}$     & F$^{adv}$  & M$^{adv}$         \\ \hline
FIM~\cite{fim}     & 29.0          & 9.3           & 5.6           & 13.8        & 6.8           & 3.6         & 2.4           \\
DI~\cite{dim}      & 46.9          & 21.7          & 14.4          & 28.0        & 12.4          & 7.9         & 6.1           \\
DFANet~\cite{dfanet}  & 50.7          & 15.5          & 12.5          & 25.6        & 11.0          & 5.8         & 3.2           \\
VMI~\cite{vt}     & 49.7          & 23.9          & 18.7          & 30.1        & 18.3          & 12.8        & 11.2          \\
SSA~\cite{ssa6}     & 55.0          & 21.9          & 24.0          & 28.8        & 14.2          & 9.0         & 6.1           \\
SIA~\cite{sia}     & 52.6          & 26.3          & 19.6          & 29.8        & 18.3          & 11.0        & 9.5           \\
BPFA~\cite{bpfa}    & 46.7          & 12.9          & 9.2           & 20.1        & 8.9           & 4.7         & 3.1           \\
BSR~\cite{bsr}     & 35.3          & 14.7          & 7.3           & 19.2        & 9.9           & 6.6         & 4.3           \\
Ours    & \textbf{99.4} & \textbf{90.3} & \textbf{96.4} & \textbf{74.0} & \textbf{69.9} & \textbf{42.0} & \textbf{57.7} \\ \hline
\end{tabular}
}
\end{table}

\begin{table}[]
\caption{Comparisons of black-box ASR (\%) results for attacks using IRSE50 as the surrogate model on the LFW dataset. I, S, F, M denote IR152, IRSE50, FaceNet, and MobileFace, respectively.}
\label{tab:compare_lfw_s}
\centering
\adjustbox{width=\columnwidth}{
\begin{tabular}{l|ccccccc}
\hline
Attacks & I         & F       & M            & I$^{adv}$    & S$^{adv}$     & F$^{adv}$  & M$^{adv}$        \\ \hline
FIM~\cite{fim}    & 32.3          & 15.5          & 79.1          & 9.8           & 17.5          & 5.5           & 5.7           \\
DI~\cite{dim}    & 59.9          & 47.5          & 97.7          & 25.9          & 41.5          & 15.6          & 23.8          \\
DFANet~\cite{dfanet} & 44.3          & 26.7          & 96.9          & 15.3          & 28.0          & 8.6           & 12.4          \\
VMI~\cite{vt}    & 54.0          & 34.0          & 96.4          & 22.5          & 37.6          & 13.1          & 20.3          \\
SSA~\cite{ssa6}    & 58.8          & 37.1          & 97.3          & 22.4          & 38.5          & 12.3          & 18.1          \\
SIA~\cite{sia}   & 58.4          & 41.4          & 98.2          & 22.2          & 37.6          & 13.9          & 23.3          \\
BPFA~\cite{bpfa}   & 54.4          & 27.5          & 94.6          & 17.6          & 29.4          & 8.1           & 12.8          \\
BSR~\cite{bsr}   & 28.7          & 18.4          & 84.5          & 9.2           & 16.3          & 6.5           & 7.6           \\
Ours    & \textbf{74.4} & \textbf{89.8} & \textbf{98.3} & \textbf{57.9} & \textbf{68.2} & \textbf{38.3} & \textbf{59.7} \\ \hline
\end{tabular}
}
\end{table}

\begin{table}[]
\caption{Comparisons of black-box ASR (\%) results for attacks using FaceNet as the surrogate model on the LFW dataset. I, S, F, M denote IR152, IRSE50, FaceNet, and MobileFace, respectively.}
\label{tab:compare_lfw_f}
\centering
\adjustbox{width=\columnwidth}{
\begin{tabular}{l|ccccccc}
\hline
Attacks & I         & S        & M            & I$^{adv}$    & S$^{adv}$     & F$^{adv}$  & M$^{adv}$         \\ \hline
FIM~\cite{fim}     & 7.8           & 12.5          & 5.4           & 7.5           & 6.9           & 17.2          & 2.5           \\
DI~\cite{dim}      & 18.6          & 32.2          & 18.5          & 18.0          & 15.8          & 30.2          & 9.9           \\
DFANet~\cite{dfanet}  & 12.1          & 22.2          & 11.7          & 11.3          & 10.4          & 25.1          & 5.5           \\
VMI~\cite{vt}     & 24.4          & 35.1          & 20.7          & 24.4          & 23.2          & 36.3          & 15.2          \\
SSA~\cite{ssa6}     & 21.6          & 44.8          & 30.8          & 17.7          & 17.0          & 31.9          & 10.9          \\
SIA~\cite{sia}     & 29.1          & 42.9          & 26.2          & 28.7          & 23.9          & 38.3          & 16.7          \\
BPFA~\cite{bpfa}    & 17.3          & 31.6          & 14.7          & 13.4          & 13.0          & 22.6          & 7.8           \\
BSR~\cite{bsr}     & 28.6          & 42.4          & 25.9          & 26.2          & 24.3          & 34.2          & 16.0          \\
Ours    & \textbf{42.6} & \textbf{65.0} & \textbf{56.9} & \textbf{47.3} & \textbf{45.4} & \textbf{54.0} & \textbf{31.1} \\ \hline
\end{tabular}
}
\end{table}

\subsection{Comparison Studies on CelebA-HQ} \label{sec:celeba_hq}
To further validate the efficacy of our proposed attack method, we create adversarial examples utilizing the CelebA-HQ dataset. The black-box performance of our approach, which employs IR152, IRSE50, FaceNet, and MobileFace as surrogate models on the CelebA-HQ dataset, is presented in \cref{tab:compare_celeba_hq_i}, \cref{tab:compare_celeba_hq_s}, \cref{tab:compare_celeba_hq_f}, and \cref{tab:compare_celeba_hq_m}, respectively. These results consistently indicate that our method outperforms the baseline attacks, thereby highlighting its effectiveness in improving the transferability of adversarial examples.

\begin{table}[]
\caption{Comparisons of black-box ASR (\%) results for attacks using IR152 as the surrogate model on the CelebA-HQ dataset. I, S, F, M denote IR152, IRSE50, FaceNet, and MobileFace, respectively.}
\label{tab:compare_celeba_hq_i}
\centering
\adjustbox{width=\columnwidth}{
\begin{tabular}{l|ccccccc}
\hline
Attacks & S       & F        & M     & I$^{adv}$     & S$^{adv}$     & F$^{adv}$     & M$^{adv}$     \\ \hline
FIM~\cite{fim}     & 39.2          & 14.1          & 12.6          & 18.3          & 12.1          & 4.3           & 4.9           \\
DI~\cite{dim}      & 57.6          & 27.6          & 27.0          & 30.1          & 20.5          & 8.5           & 10.8          \\
DFANet~\cite{dfanet}  & 61.2          & 21.5          & 22.4          & 26.6          & 17.0          & 5.7           & 7.9           \\
VMI~\cite{vt}     & 56.9          & 26.4          & 27.7          & 32.4          & 26.1          & 12.6          & 16.6          \\
SSA~\cite{ssa6}     & 62.1          & 23.3          & 30.7          & 30.6          & 19.4          & 9.1           & 10.3          \\
SIA~\cite{sia}     & 60.8          & 26.9          & 30.4          & 30.4          & 24.3          & 12            & 14.2          \\
BPFA~\cite{bpfa}    & 54.6          & 15.8          & 16.4          & 22.0          & 14.2          & 5.0             & 5.4           \\
BSR~\cite{bsr}     & 42.9          & 17.3          & 15.0          & 21.1          & 15.4          & 6.6           & 6.8           \\
Ours    & \textbf{98.0} & \textbf{82.4} & \textbf{95.1} & \textbf{53.5} & \textbf{61.5} & \textbf{27.6} & \textbf{52.8} \\ \hline
\end{tabular}
}
\end{table}

\begin{table}[]
\caption{Comparisons of black-box ASR (\%) results for attacks using IRSE50 as the surrogate model on the CelebA-HQ dataset. I, S, F, M denote IR152, IRSE50, FaceNet, and MobileFace, respectively.}
\label{tab:compare_celeba_hq_s}
\centering
\adjustbox{width=\columnwidth}{
\begin{tabular}{l|ccccccc}
\hline
Attacks & I             & F             & M             & I$^{adv}$     & S$^{adv}$     & F$^{adv}$     & M$^{adv}$     \\ \hline
FIM~\cite{fim}     & 36.3          & 16.2          & 80.5          & 10.5          & 21.6          & 5.1           & 9.5           \\
DI~\cite{dim}      & 59.3          & 42.8          & 97.6          & 20.5          & 39.4          & 12.0          & 28.5          \\
DFANet~\cite{dfanet}  & 45.9          & 25.5          & 97.0          & 14.0          & 30.8          & 6.6           & 15.4          \\
VMI~\cite{vt}     & 56.7          & 31.9          & 96.6          & 18.5          & 37.5          & 9.9           & 24.3          \\
SSA~\cite{ssa6}     & 58.3          & 34.8          & 97.5          & 19.1          & 38.2          & 8.9           & 22.1          \\
SIA~\cite{sia}     & 60.7          & 40.0          & 97.9          & 20.1          & 40.1          & 11.5          & 26.4          \\
BPFA~\cite{bpfa}    & 56.8          & 27.9          & 95.3          & 16.7          & 32.8          & 7.5           & 17.3          \\
BSR~\cite{bsr}     & 35.7          & 19.9          & 86.4          & 11.2          & 20.4          & 5.3           & 12.0          \\
Ours    & \textbf{68.9} & \textbf{81.7} & \textbf{98.0} & \textbf{40.2} & \textbf{60.6} & \textbf{25.5} & \textbf{53.8} \\ \hline
\end{tabular}
}
\end{table}

\begin{table}[]
\caption{Comparisons of black-box ASR (\%) results for attacks using FaceNet as the surrogate model on the CelebA-HQ dataset. I, S, F, M denote IR152, IRSE50, FaceNet, and MobileFace, respectively.}
\label{tab:compare_celeba_hq_f}
\centering
\adjustbox{width=\columnwidth}{
\begin{tabular}{l|ccccccc}
\hline
Attacks & I             & S             & M             & I$^{adv}$     & S$^{adv}$     & F$^{adv}$     & M$^{adv}$     \\ \hline
FIM~\cite{fim}     & 10.7          & 16.5          & 9.6           & 7.2           & 8.2           & 13.0          & 4.4           \\
DI~\cite{dim}      & 22.8          & 30.4          & 22.9          & 15.0          & 19.9          & 21.7          & 11.8          \\
DFANet~\cite{dfanet}  & 15.2          & 24.3          & 19.7          & 10.0          & 14.1          & 18.0          & 7.7           \\
VMI~\cite{vt}     & 26.4          & 36.4          & 25.7          & 19.5          & 25.2          & 25.3          & 16.4          \\
SSA~\cite{ssa6}     & 23.5          & 41.5          & 36.1          & 14.9          & 19.4          & 22.4          & 13.2          \\
SIA~\cite{sia}     & 29.3          & 44.5          & 35.7          & 21.5          & 26.0          & 27.6          & 18.4          \\
BPFA~\cite{bpfa}    & 20.0          & 31.0          & 21.7          & 12.1          & 14.0          & 16.8          & 7.9           \\
BSR~\cite{bsr}     & 27.8          & 43.9          & 34.0          & 19.2          & 25.9          & 26.0          & 18.3          \\
Ours    & \textbf{35.5} & \textbf{56.9} & \textbf{58.7} & \textbf{29.9} & \textbf{36.2} & \textbf{32.2} & \textbf{28.3} \\ \hline
\end{tabular}
}
\end{table}

\begin{table}[]
\caption{Comparisons of black-box ASR (\%) results for attacks using MobileFace as the surrogate model on the CelebA-HQ dataset. I, S, F, M denote IR152, IRSE50, FaceNet, and MobileFace, respectively.}
\label{tab:compare_celeba_hq_m}
\centering
\adjustbox{width=\columnwidth}{
\begin{tabular}{l|ccccccc}
\hline
Attacks & I         & S        & F       & I$^{adv}$     & S$^{adv}$     & F$^{adv}$     & M$^{adv}$     \\ \hline
FIM~\cite{fim}     & 7.2           & 69.0          & 8.0           & 3.3           & 5.4           & 2.4           & 12.0          \\
DI~\cite{dim}      & 25.0          & 97.0          & 32.0          & 11.0          & 21.4          & 8.2           & 39.0          \\
DFANet~\cite{dfanet}  & 10.7          & 85.0          & 12.6          & 3.7           & 9.1           & 2.9           & 18.8          \\
VMI~\cite{vt}     & 20.2          & 94.7          & 19.6          & 7.8           & 15.8          & 4.4           & 31.5          \\
SSA~\cite{ssa6}     & 22.0          & 95.4          & 21.0          & 7.9           & 15.2          & 5.3           & 33.1          \\
SIA~\cite{sia}     & 25.4          & 96.6          & 25.9          & 8.9           & 18.4          & 5.9           & 35.8          \\
BPFA~\cite{bpfa}   & 20.7          & 94.7          & 17.5          & 6.7           & 13.8          & 4.5           & 29.7          \\
BSR~\cite{bsr}     & 10.8          & 77.1          & 11.9          & 3.8           & 8.1           & 2.8           & 15.1          \\
Ours    & \textbf{59.6} & \textbf{97.2} & \textbf{83.5} & \textbf{37.1} & \textbf{57.5} & \textbf{27.4} & \textbf{61.7} \\ \hline
\end{tabular}
}
\end{table}

\begin{figure}[t]
	\begin{center}
		\centerline{\includegraphics[width=\columnwidth]{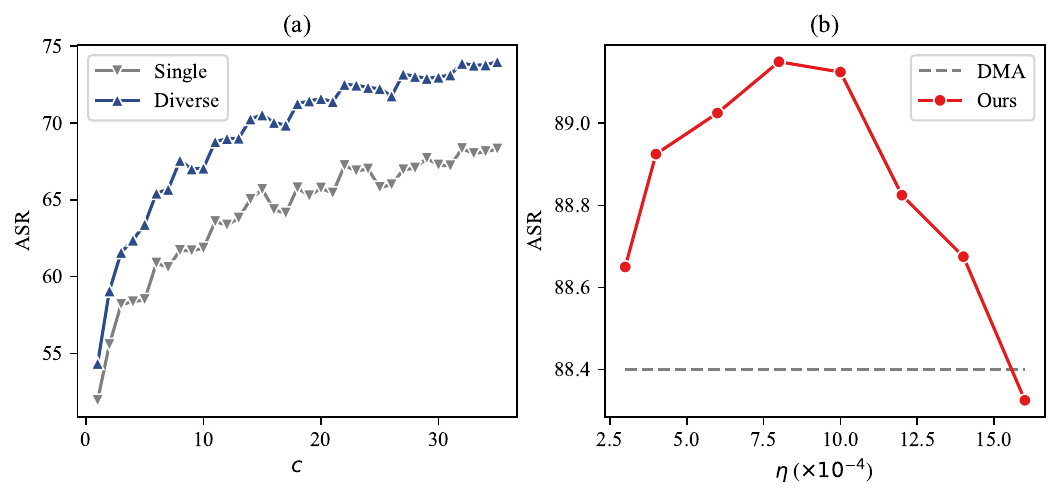}}
		\caption{The hyper-parameter analysis on the (a) $c$ and (b) $\eta$.
		}
		\label{fig:m_and_step}
	\end{center}
    \vspace{-0.8cm}
\end{figure}

\subsection{Hyper-parameter Analysis Studies}\label{sec:hyper_sen_study}
\textbf{The hyper-parameter analysis on the $c$ value.}
The value of $c$ determines the number of ensembles in our proposed attack method, which significantly affects its performance. Hence, we conduct ablation studies on $c$ using the LFW dataset with MobileFace as the surrogate model. To further verify the effectiveness of diverse parameters in enhancing transferability, we select two types of attack methods for comparison. Firstly, we use MobileFace models fine-tuned by a pre-trained backbone and a randomly initialized head in each epoch to generate adversarial examples. We term this adversarial attack method `Single'. Secondly, we choose the models trained by `Single' and MobileFace models trained by a randomly initialized backbone and head in each epoch to create adversarial examples, implying that the parameters of the trained models are more diverse. We term this attack method `Diverse'. The average ASR on IR152, IRSE50, FaceNet, and MobileFace is demonstrated in the left plot of \cref{fig:m_and_step}. The left plot of \cref{fig:m_and_step} shows that the ASR increases and then converges as $c$ increases. To analyze the reason, we need to consider the property of $c$. $c$ determines the number of models to be aggregated. If more models are aggregated in each training epoch, the aggregation capacity will increase. If $c$ continues to increase, due to the similarity of the aggregated models in the later epochs, the ASR converges. Moreover, the left plot of Figure \ref{fig:m_and_step} demonstrates that the performance of `Diverse' is higher than that of `Single', which verifies the effectiveness of parameter diversity in improving transferability of crafted adversarial examples.

\noindent \textbf{The hyper-parameter analysis on the $\eta$ value.}
The $\eta$ value is the step size of beneficial perturbations, which is a key hyperparameter in our proposed attack method. We will conduct ablation studies on this parameter using the LFW dataset with MobileFace as the surrogate model. The average ASR on IR152, IRSE50, FaceNet, and MobileFace is shown in the right plot of \cref{fig:m_and_step}. To assess the effectiveness of hard models in enhancing the transferability of adversarial examples, we use the Diverse Model Aggregation (DMA) as a baseline for comparison. DMA replaces the hard models in our method with their corresponding vanilla models. From the right plot of \cref{fig:m_and_step}, we observe that as the step size of beneficial perturbations increases, the ASR initially rises and then declines.
To understand this behavior, we should consider the nature of beneficial perturbations. These perturbations are added to the feature maps of FR models to increase loss when crafting adversarial examples, effectively transforming FR models into hard models. Increasing the step size initially boosts transferability by strengthening the transition to hard models. However, further increasing the step size can degrade the features in the feature maps during forward propagation, ultimately reducing overall attack performance.
Additionally, the right plot of \cref{fig:m_and_step} demonstrates that the optimal performance of our proposed method surpasses that of DMA, further validating the effectiveness of the hard model ensemble in our attack method.

\begin{figure}[]
	\begin{center}
		\centerline{\includegraphics[width=80mm]{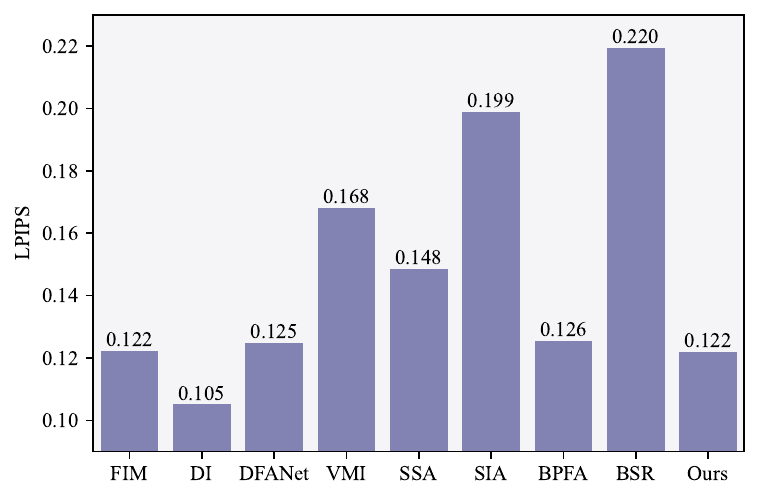}}
		\caption{Comparison of LPIPS values across various attacks, with lower values signifying superior visual quality.
		}
		\label{fig:visual_quality_study}
	\end{center}
    \vspace{-0.8cm}
\end{figure}

\subsection{Visual Quality Study} \label{sec:visual_quality}
\label{sec:visual_quality}
Furthermore, we evaluate the visual quality of our proposed method against that of previous attack methods. We choose FIM, DI, DFANet, VMI, SSA, SIA, BPFA, and BSR as comparative baselines and generate adversarial examples using MobileFace as the surrogate model on the LFW dataset. The experimental configuration is consistent with the one detailed in \cref{tab:compare_lfw_m}. The outcomes are depicted in \cref{fig:visual_quality_study}. As shown in \cref{fig:visual_quality_study}, our proposed method achieves visual quality performance on par with other methods. Notably, the transferability of the adversarial examples generated by our method significantly exceeds that of the baselines, which further underscores the superiority of our proposed method.

\subsection{Ethics and Potential Broader Impact} \label{sec:broader_impact}
This paper introduces research that contributes to the advancement of the field within Computer Vision and Pattern Recognition. The attack method we propose poses a potential threat to the security of FR models. Our goal is to heighten awareness through this proposed method and strengthen the resilience of FR models against such vulnerabilities. We prohibit the use of the attack method presented in this paper for commercial purposes. Furthermore, we explicitly forbid the application of the proposed attack method in any attempt to compromise commercial platforms for personal gain.

\end{document}